\def\year{2020}\relax
\documentclass[letterpaper]{article} % DO NOT CHANGE THIS
\usepackage{aaai20}  % DO NOT CHANGE THIS
\usepackage{times}  % DO NOT CHANGE THIS
\usepackage{helvet} % DO NOT CHANGE THIS
\usepackage{courier}  % DO NOT CHANGE THIS
\usepackage[hyphens]{url}  % DO NOT CHANGE THIS
\usepackage{graphicx} % DO NOT CHANGE THIS
\usepackage{color} % DO NOT CHANGE THIS
\def\UrlFont{\rm}  % DO NOT CHANGE THIS
\usepackage{graphicx}  % DO NOT CHANGE THIS
\usepackage{color}
\usepackage{soul}
\usepackage{url}

\usepackage[utf8]{inputenc}
\usepackage{mathtools}
\usepackage{algorithm}
\usepackage{algorithmic}
\usepackage{esvect}
\usepackage[toc,page]{appendix}
\usepackage{comment}

\usepackage{amsmath}
\DeclareMathOperator*{\argmax}{arg\,max}

\DeclarePairedDelimiter{\norm}{\lVert}{\rVert}

\title{Discovering and Categorising Language Biases in Reddit\thanks{Author's copy of the paper accepted at the International AAAI Conference on Web and Social Media (ICWSM 2021).}}
\author{Xavier Ferrer\textsuperscript{\rm +},
Tom van Nuenen\textsuperscript{\rm +},
Jose M. Such\textsuperscript{\rm +} and 
Natalia Criado\textsuperscript{\rm +}\\
\textsuperscript{\rm +} \Large Department of Informatics, King's College London\\ 
\{xavier.ferrer\_aran, tom.van\_nuenen, jose.such, natalia.criado\}@kcl.ac.uk
}

\begin{document}

\maketitle

\begin{abstract}
We present a data-driven approach using word embeddings to discover and categorise language biases on the discussion platform Reddit. As spaces for isolated user communities, platforms such as Reddit are increasingly connected to issues of racism, sexism and other forms of discrimination. Hence, there is a need to monitor the language of these groups. One of the most promising AI approaches to trace linguistic biases in large textual datasets involves word embeddings, which transform text into high-dimensional dense vectors and capture semantic relations between words. Yet, previous studies require predefined sets of potential biases to study, e.g., whether gender is more or less associated with particular types of jobs. This makes these approaches unfit to deal with smaller and community-centric datasets such as those on Reddit, which contain smaller vocabularies and slang, as well as biases that may be particular to that community. This paper proposes a data-driven approach to automatically discover language biases encoded in the vocabulary of online discourse communities on Reddit. In our approach, protected attributes are connected to evaluative words found in the data, which are then categorised through a semantic analysis system. We verify the effectiveness of our method by comparing the biases we discover in the Google News dataset with those found in previous literature. We then successfully discover gender bias, religion bias, and ethnic bias in different Reddit communities. We conclude by discussing potential application scenarios and limitations of this data-driven bias discovery method.
\end{abstract}

%%%%%%%%%%%%%%%%%%%%%%%
%
%  Introduction
% 
%%%%%%%%%%%%%%%%%%%%%%%

\section{Introduction}\label{sec:intro}

This paper proposes a general and data-driven approach to discovering linguistic biases towards protected attributes, such as gender, in online communities. Through the use of word embeddings and the ranking and clustering of biased words, we discover and categorise biases in several English-speaking communities on Reddit, using these communities' own forms of expression.

Reddit is a web platform for social news aggregation, web content rating, and discussion. It serves as a platform for multiple, linked topical discussion forums, as well as a network for shared identity-making \cite{Papacharissi2015}. Members can submit content such as text posts, pictures, or direct links, which is organised in distinct message boards curated by interest communities. These `subreddits' are distinct message boards curated around particular topics, such as /r/pics for sharing pictures or /r/funny for posting jokes\footnote{Subreddits are commonly spelled with the prefix `/r/'.}. Contributions are submitted to one specific subreddit, where they are aggregated with others. 

Not least because of its topical infrastructure, Reddit has been a popular site for Natural Language Processing studies – for instance, to successfully classify mental health discourses \cite{Balani2015}, and domestic abuse stories \cite{Schrading2015}. % and to explore social norms \cite{Kasunic2018}. 
LaViolette and Hogan have recently augmented traditional NLP and machine learning techniques with platform metadata, allowing them to interpret misogynistic discourses in different subreddits \cite{LaViolette2019}. Their focus on discriminatory language is mirrored in other studies, which have pointed out the propagation of sexism, racism, and `toxic technocultures' on Reddit using a combination of NLP and discourse analysis \cite{Mountford2018}. What these studies show is that social media platforms such as Reddit not merely reflect a distinct offline world, but increasingly serve as constitutive spaces for contemporary ideological groups and processes. 

Such ideologies and biases become especially pernicious when they concern vulnerable groups of people that share certain \emph{protected attributes} – including ethnicity, gender, and religion \cite{Grgic-Hlaca2018}. Identifying language biases towards these protected attributes can offer important cues to tracing harmful beliefs fostered in online spaces. Recently, NLP research using word embeddings has been able to do just that \cite{Caliskan2017,garg2018word}. However, due to the reliance on predefined concepts to formalise bias, these studies generally make use of larger textual corpora, such as the widely used Google News dataset \cite{mikolov2013distributed}. This makes these methods less applicable to social media platforms such as Reddit, as communities on the platform tend to use language that operates within conventions defined by the social group itself. Due to their topical organisation, subreddits can be thought of as `discourse communities' \cite{Kehus2010}, which generally have a broadly agreed set of common public goals and functioning mechanisms of intercommunication among its members. They also share discursive expectations, as well as a specific lexis \cite{Swales2011}. As such, they may carry biases and stereotypes that do not necessarily match those of society at large. At worst, they may constitute cases of hate speech, 'language that is used to expresses hatred towards a targeted group or is intended to be derogatory, to humiliate, or to insult the members of the group' \cite{Davidson2017}. The question, then, is how to discover the biases and stereotypes associated with protected attributes that manifest in particular subreddits – and, crucially, which linguistic form they take.

This paper aims to bridge NLP research in social media, which thus far has not connected discriminatory language to protected attributes, and research tracing language biases using word embeddings. Our contribution consists of developing a general approach to \textit{discover} and \textit{categorise} biased language towards protected attributes in Reddit communities. We use word embeddings to determine the most biased words towards protected attributes, apply k-means clustering combined with a semantic analysis system to label the clusters, and use sentiment polarity to further specify biased words. We validate our approach with the widely used Google News dataset before applying it to several Reddit communities. In particular, we identified and categorised gender biases in /r/TheRedPill and /r/dating\_advice, religion biases in /r/atheism and ethnicity biases in /r/The\_Donald.

%%%%%%%%%%%%%%%%%%%%%%%
%
%  Related work
% 
%%%%%%%%%%%%%%%%%%%%%%%

\section{Related work}\label{sec:validation}

Linguistic biases have been the focus of language analysis for quite some time \cite{Wetherell1992,Holmes2008,garg2018word,bhatia2017semantic}.
Language, it is often pointed out, functions as both a reflection and perpetuation of stereotypes that people carry with them. Stereotypes can be understood as ideas about how (groups of) people commonly behave \cite{VanMiltenburg2016}. As cognitive constructs, they are closely related to essentialist beliefs: the idea that members of some social category share a deep, underlying, inherent nature or `essence', causing them to be fundamentally similar to one another and across situations \cite{Carnaghi2008}. One form of linguistic behaviour that results from these mental processes is that of linguistic bias: `a systematic asymmetry in word choice as a function of the social category to which the target belongs.' \cite[p.313]{Beukeboom2014}. 

The task of tracing linguistic bias is accommodated by recent advances in AI \cite{aran2019attesting}. One of the most promising approaches to trace biases is through a focus on the distribution of words and their similarities in word embedding modelling. The encoding of language in word embeddings answers to the distributional hypothesis in linguistics, which holds that the statistical contexts of words capture much of what we mean by meaning \cite{Sahlgren2008}. In word embedding models, each word in a given dataset is assigned to a high-dimensional vector such that the geometry of the vectors captures semantic relations between the words – e.g. vectors being closer together correspond to distributionally similar words \cite{Collobert2011}. In order to capture accurate semantic relations between words, these models are typically trained on large corpora of text. One example is the Google News word2vec model, a word embeddings model trained on the Google News dataset \cite{mikolov2013distributed}. 

Recently, several studies have shown that word embeddings are strikingly good at capturing human biases in large corpora of texts found both online and offline \cite{bolukbasi2016man,Caliskan2017,VanMiltenburg2016}. In particular, word embeddings approaches have proved successful in creating analogies \cite{bolukbasi2016man}, and quantifying well-known societal biases and stereotypes \cite{Caliskan2017,garg2018word}. These approaches test for predefined biases and stereotypes related to protected attributes, e.g., for gender, that males are more associated with a professional career and females with family. In order to define sets of words capturing potential biases, which we call `evaluation sets', previous studies have taken word sets from Implicit Association Tests (IAT) used in social psychology. This test detects the strength of a person's automatic association between mental representations of objects in memory, in order to assess bias in general societal attitudes. \cite{greenwald1998measuring}. The evaluation sets yielded from IATs are then related to ontological concepts representing protected attributes, formalised as a `target set'. This means two supervised word lists are required; e.g., the protected attribute `gender' is defined by target words related to men (such as \{`he', `son', `brother', \ldots\}) and women (\{`she', `daughter', `sister', ...\}), and potential biased concepts are defined in terms of sets of evaluative terms largely composed of adjectives, such `weak' or `strong'. Bias is then tested through the positive relationship between these two word lists. Using this approach, Caliskan et al. were able to replicate IAT findings by introducing their Word-Embedding Association Test (WEAT). The cosine similarity between a pair of vectors in a word embeddings model proved analogous to reaction time in IATs, allowing the authors to determine biases between target and evaluative sets. The authors consider such bias to be `stereotyped' when it relates to aspects of human culture known to lead to harmful behaviour \cite{Caliskan2017}. 

Caliskan et al. further demonstrate that word embeddings can capture imprints of historic biases, ranging from morally neutral ones (e.g. towards insects) to problematic ones (e.g. towards race or gender) \cite{Caliskan2017}. For example, in a gender-biased dataset, the vector for adjective `honourable' would be closer to the vector for the `male' gender, whereas the vector for `submissive' would be closer to the `female' gender. Building on this insight, Garg et.al. have recently built a framework for a diachronic analysis of word embeddings, which they show incorporate changing `cultural stereotypes' \cite{garg2018word}. The authors demonstrate, for instance, that during the second US feminist wave in the 1960, the perspectives on women as portrayed in the Google News dataset fundamentally changed. More recently, WEAT was also adapted to BERT embeddings \cite{kurita2019measuring}.

What these previous approaches have in common is a reliance on predefined evaluative word sets, which are then tested on target concepts that refer to protected attributes such as gender. This makes it difficult to transfer these approaches to other – and especially smaller – linguistic datasets, which do not necessarily include the same vocabulary as the evaluation sets. Moreover, these tests are only useful to determine  predefined biases for predefined concepts. Both of these issues are relevant for the subreddits we are interested in analysing here: they are relatively small, are populated by specific groups of users, revolve around very particular topics and social goals, and often involve specialised vocabularies. The biases they carry, further, are not necessarily representative of broad `cultural stereotypes'; in fact, they can be antithetical even to common beliefs. An example in /r/TheRedPill, one of our datasets, is that men in contemporary society are oppressed by women \cite{Marwick2017}. Within the transitory format of the online forum, these biases can be linguistically negotiated – and potentially transformed – in unexpected ways. 

Hence, while we have certain ideas about which kinds of protected attributes to expect biases against in a community (e.g. gender biases in /r/TheRedPill), it is hard to tell in advance which concepts will be associated to these protected attributes, or what linguistic form biases will take. The approach we propose extracts and aggregates the  words relevant within each subreddit in order to identify biases regarding protected attributes as they are encoded in the linguistic forms chosen by the community itself.

%%%%%%%%%%%%%%%%%%%%%%%
%
%  OVERVIEW OF OUR FRAMEWORK
% 
%%%%%%%%%%%%%%%%%%%%%%%

\section{Discovering language biases}\label{sec:overview}

In this section we present our approach to discover linguistic biases. 

\subsection{Most biased words}

Given a word embeddings model of a corpus (for instance, trained with textual comments from a Reddit community)  and two sets of target words representing two concepts we want to compare and discover biases from, we identify the most biased words towards these concepts in the community. Let $S_1 = \{w_{i}, w_{i+1}, ..., w_{i+n}\}$ be a set of target words $w$ related to a concept (e.g \emph{\{he, son, his, him, father}, and \emph{male\}} for concept \emph{male}), and $\vv{c_1}$ the centroid of $S_1$ estimated by averaging the embedding vectors of word $w \in S_1$. Similarly, let $S_2 = \{w_{j}, w_{j+1}, ..., w_{j+m}\}$ be a second set of target words with centroid  $\vv{c_2}$ (e.g. \emph{\{she, daughter, her, mother}, and \emph{female\}} for concept \emph{female}). A word $w$ is biased towards $S_1$ with respect to $S_2$ when the cosine similarity\footnote{Alternative bias definitions are possible here, such as the \emph{direct bias} measure defined in \cite{bolukbasi2016man}. In fact, when compared experimentally with our metric in r/TheRedPill, we obtain a Jaccard index of 0.857 (for female gender) and 0.864 (for male) regarding the list of 300 most-biased adjectives generated with the two bias metrics. Similar results could also be obtained using the relative norm bias metric, as shown in \cite{garg2018word}.} between the embedding of $\vv{w}$, is higher for $\vv{c_1}$ than for $\vv{c_2}$.

\begin{equation}
Bias(w, c_1, c_2) = cos(\vv{w}, \vv{c_1} ) - cos( \vv{w}, \vv{c_2})
\label{eq:mb}
\end{equation}
\noindent where $cos(u, v) =  \frac{u \cdot v}{ \norm{u}_2 \norm{v}_2 } $.
Positive values of $Bias$ mean a word $w$ is more  biased towards $S_1$, while negative values of $Bias$ means $w$ is more biased towards $S_2$. 

Let $V$ be the vocabulary of a word embeddings model. We identify the $k$ most biased words towards $S_1$ with respect to $S_2$ by ranking the words in the vocabulary $V$ using $Bias$ function from Equation \ref{eq:mb2}:

\begin{equation}
MostBiased(V, c_1, c_2) = \argmax_{w\in V} Bias( w, c_1, c_2)
\label{eq:mb2}
\end{equation}

Researchers typically focus on discovering biases and stereotypes by exploring the most biased adjectives and nouns towards two sets of target words (e.g. \emph{female} and \emph{male}). Adjectives are particularly interesting since they modify nouns by limiting, qualifying, or specifying their properties, and are often normatively charged. Adjectives carry polarity, and thus often yield more interesting insights about the type of discourses. 
In order to determine the part-of-speech (POS) of a word, we use the \texttt{nltk}\footnote{https:\/\/www.nltk.org} python library. 
POS filtering helps us removing non-interesting words in some communities such as acronyms, articles and proper names (cf. Appendix A for a performance evaluation of POS using the nltk library in the datasets used in this paper).

\begin{figure}[h]
  \centering
  \includegraphics[height=0.5\linewidth]{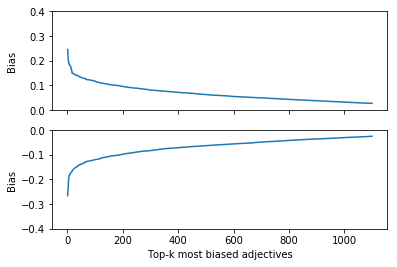}
  \caption{Bias distribution in adjectives in the r/TheRedPill}
  \label{fig:trp_qbiases}
\end{figure}

Given a vocabulary and two sets of target words (such as those for \emph{women} and \emph{men}), we rank the words from least to most biased using Equation \ref{eq:mb2}. As such, we obtain two ordered lists of the most biased words towards each target set, obtaining an overall view of the bias distribution in that particular community with respect to those two target sets. 
For instance, Figure \ref{fig:trp_qbiases} shows the bias distribution of words towards women (top) and men (bottom) target sets in /r/TheRedPill.
Based on the distribution of biases towards each target set in each subreddit, we determine the threshold of how many words to analyse by selecting the top words using Equation \ref{eq:mb2}.
All targets sets used in our work are compiled from previous experiments (listed in Appendix \ref{ap:targetsets}).

\subsection{Sentiment Analysis}\label{sec:sa}
To further specify the biases we encounter, we take the sentiment polarity of biased words into account. Discovering consistently strong negative polarities among the most biased words towards a target set might be indicative of strong biases, and even stereotypes, towards that specific population\footnote{Note that potentially discriminatory biases can also be encoded in a-priori sentiment-neutral words. The fact that a word is not tagged with a negative sentiment does not exclude it from being discriminatory in certain contexts.}. We are interested in assessing whether the most biased words towards a population carry negative connotations, and we do so by performing a sentiment analysis over the most biased words towards each target using the \texttt{nltk} sentiment analysis python library \cite{hutto2014vader}\footnote{Other sentiment analysis tools could be used but some might return biased analyses \cite{kiritchenko2018examining}.}. We estimate the average of the sentiment of a set of words $W$ as such:
\begin{equation}
    Sent(W) = \frac{1}{|W|} \sum_{w \in W} SA(w)
    \label{eq:sav}
\end{equation}
\noindent where $SA$ returns a value $\in [-1, 1]$ corresponding to the polarity determined by the sentiment analysis system, -1 being strongly negative and +1 strongly positive. As such, $Sent(W)$ always returns a value $\in [-1,1]$.

Similarly to POS tagging, the polarity of a word depends on the context in which the word is found. Unfortunately, contextual information is not encoded in the pre-trained word embedding models commonly used in the literature. As such, we can only leverage the prior sentiment polarity of words without considering the context of the sentence in which they were used. Nevertheless, a consistent tendency towards strongly polarised negative (or positive) words can give some information about tendencies and biases towards a target set.

%%%%%%%%%%%%%%%%%%%%%%%%%%%%%%%%%%%%%%%%%%%%%
%
%   K -means + clustering
%
%%%%%%%%%%%%%%%%%%%%%%%%%%%%%%%%%%%%%%%%%%%%%
\subsection{Categorising Biases}\label{sec:discoveringstereotypes}

As noted, we aim to discover the most biased terms towards a target set. However, even when knowing those most biased terms and their polarity, considering each of them as a separate unit may not suffice in order to discover the relevant concepts they represent and, hence, the contextual meaning of the bias. Therefore, we also combine semantically related terms under broader rubrics in order to facilitate the comprehension of a community's biases.
A side benefit is that identifying concepts as a cluster of terms, instead of using individual terms, helps us tackle stability issues associated with individual word usage in word embeddings \cite{Antoniak2018} - discussed in Section \ref{sec:trp}.

We aggregate the most similar word embeddings into clusters using the well-known k-means clustering algorithm. In k-means clustering, the parameter $k$ defines the quantity of clusters into which the space will be partitioned. Equivalently, we use the \emph{reduction factor} $\in (0,1)$, $r = \frac{k}{|V|}$, where $|V|$ is the size of the vocabulary to be partitioned. 
The lower the value of $r$, the lower the quantity of clusters and their average intra-similarity, estimated by assessing the average similarity between all words in a cluster for all clusters in a partition. On the other hand, when $r$ is close to 1, we obtain more clusters and a higher cluster intra-similarity, up to when $r = 1$ where we have $|V|$ clusters of size 1, with an average intra-similarity of 1 (see Appendix \ref{ap:sec_granularity}).

In order to assign a label to each cluster, which facilitates the categorisation of biases related to each target set, we use the UCREL Semantic Analysis System (USAS)\footnote{\url{http://ucrel.lancs.ac.uk/usas/}, accessed Apr 2020}. USAS is a framework for the automatic semantic analysis and tagging of text, originally based on Tom McArthur's Longman Lexicon of Contemporary English \cite{summers1995longman}. It has a multi-tier structure with 21 major discourse fields subdivided in more fine-grained categories such as \emph{People}, \emph{Relationships} or \emph{Power}.
USAS has been extensively used for tasks such as the automatic content analysis of spoken discourse \cite{wilson1993automatic} or as a translator assistant \cite{sharoff2006assist}. The creators also offer an interactive tool\footnote{\url{http://ucrel-api.lancaster.ac.uk/usas/tagger.html}, accessed Apr 2020} to automatically tag each word in a given sentence with a USAS semantic label. 

Using the USAS system, every cluster is labelled with the most frequent tag (or tags) among the words clustered in the k-means cluster. For instance, \emph{Relationship: Intimate/sexual} and \emph{Power, organising} are two of the most common labels assigned to the gender-biased clusters of /r/TheRedPill (see Section \ref{sec:trp}). 
However, since many of the communities we explore make use of non-standard vocabularies, dialects, \emph{slang} words and grammatical particularities, the USAS automatic analysis system has occasional difficulties during the tagging process. Slang and community-specific words such as \emph{dateable} (someone who is \emph{good enough} for dating) or \emph{fugly} (used to describe someone considered very ugly) are often left uncategorised. In these cases, the uncategorised clusters receive the label (or labels) of the most similar cluster in the partition, determined by analysing the cluster centroid distance between the unlabelled cluster and the other cluster centroids in the partition.
For instance, in /r/TheRedPill, the cluster \emph{(interracial)} (a one-word cluster) was initially left unlabelled. The label was then updated to \emph{Relationship: Intimate/sexual} after copying the label of the most similar cluster, which was \emph{(lesbian, bisexual)}. 

Once all clusters of the partition are labelled, we rank all labels for each target based on the quantity of clusters tagged and, in case of a tie, based on the quantity of words of the clusters tagged with the label. By comparing the rank of the labels between the two target sets and combining it with an analysis of the clusters' average polarities, we obtain a general understanding of the most frequent conceptual biases towards each target set in that community. We particularly focus on the most relevant clusters based on rank difference between target sets or other relevant characteristics such as average sentiment of the clusters, but we also include the top-10 most frequent conceptual biases for each dataset (Appendix \ref{ap:targetsets}).

%%%%%%%%%%%%%%%%
%
% Google News 
%
%%%%%%%%%%%%%%%%
\section{Validation on Google News}\label{sec:google_news2}
In this section we use our approach to discover gender biases in the Google News pre-trained model\footnote{We used the Google news model (\url{https://code.google.com/archive/p/word2vec/}), due to its wide usage in relevant literature. However, our method could also be extended and applied in newer embedding models such as ELMO and BERT.}, and compare them with previous findings \cite{garg2018word,Caliskan2017} to prove that our method yields relevant results that complement those found in the existing literature. 

The Google News embedding model contains 300-dimensional vectors for 3 million words and phrases, trained on part of the US Google News dataset containing about 100 billion words. Previous research on this model reported gender biases among others \cite{garg2018word}, and we repeated the three WEAT experiments related to gender from \cite{Caliskan2017} in Google News. These WEAT experiments compare the association between male and female target sets to evaluative sets indicative of gender binarism, including \emph{career} Vs \emph{family}, \emph{math} Vs \emph{arts}, and \emph{science} Vs \emph{arts}, where the first sets include a-priori male-biased words, and the second include female-biased words (see  Appendix \ref{ap:targetsets}). In all three cases, the WEAT tests show significant $p$-values ($p=10^{-3}$ for \emph{career/family},  $p=0.018$ for \emph{math/arts}, and $p=10^{-2}$ for \emph{science/arts}), indicating relevant gender biases with respect to the particular word sets.

Next, we use our approach on the Google News dataset to discover the gender biases of the community, and to identify whether the set of conceptual biases and USAS labels confirms the findings of previous studies with respect to arts, science, career and family. 

For this task, we follow the method stated in Section~\ref{sec:overview} and start by observing the bias distribution of the dataset, in which we identify the 5000 most biased uni-gram adjectives and nouns towards 'female' and `male' target sets. The experiment is performed with a reduction factor $r=0.15$, although this value could be modified to zoom out/in the different clusters (see Appendix \ref{ap:sec_granularity}). 
After selecting the most biased nouns and adjectives, the k-means clustering partitioned the resulting vocabulary in 750 clusters for women and man.
There is no relevant average prior sentiment difference between male and female-biased clusters. 

Table \ref{tb:gnews_comparison2} shows some of the most relevant labels used to tag the female and male-biased clusters in the Google News dataset, where $R. Female$ and $R. Male$ indicate the rank importance of each label among the sets of labels used to tag each cluster for each gender. Character `-' indicates that the label is not found among the labels biased towards the target set. Due to space limitations, we only report the most pronounced biases based on frequency and rank difference between target sets (see Appendix \ref{ap:sec_extended} for the rest top-ten labels). 
Among the most frequent concepts more biased towards women, we find labels such as \emph{Clothes and personal belongings}, \emph{People: Female}, \emph{Anatomy and physiology}, and \emph{Judgement of appearance (pretty etc.)}. 
In contrast, labels related to strength and power, such as \emph{Warfare, defence and the army; weapons}, \emph{Power, organizing}, followed by \emph{Sports}, and \emph{Crime}, are among the most frequent concepts much more biased towards men.

\begin{table}[]
\scriptsize
\centering
\caption{Google News most relevant cluster labels (gender). }
\begin{tabular}{| l | r r |} \hline
Cluster Label  & R. Female & R. Male  \\ \hline
\multicolumn{3}{c}{Relevant to \emph{Female}} \\ \hline
Clothes and personal belongings                    	&	     3 	&	    20           \\
People: Female                                    	&	     4 	&	     -           \\
Anatomy and physiology                             	&	     5 	&	    11           \\
Cleaning and personal care                         	&	     7 	&	    68           \\
Judgement of appearance (pretty etc.)              	&	     9 	&	    29           \\
\hline
\multicolumn{3}{c}{Relevant to \emph{Male}} \\ \hline
Warfare, defence and the army; weapons              &	     - 	&	     3           \\
Power, organizing                                   &	     8 	&	     4           \\
Sports                                             	&	     - 	&	     7           \\
Crime                                              	&	     68 	&	    8           \\
Groups and affiliation                             	&	     - 	&	     9           \\
\hline
\end{tabular}
\label{tb:gnews_comparison2}
\end{table}

We now compare with the biases that had been tested in prior works by, first, mapping the USAS labels related to \emph{career, family, arts, science} and \emph{maths} based on an analysis of the WEAT word sets and the category descriptions provided in the USAS website (see Appendix \ref{ap:targetsets}), and second, evaluating how frequent are those labels among the set of most biased words towards women and men. 
The USAS labels related to \emph{career} are more frequently biased towards men, with a total of 24 and 38 clusters for women and men, respectively, containing words such as `barmaid' and `secretarial' (for women) and `manager' (for men). \emph{Family}-related clusters are strongly biased towards women, with twice as many clusters for women (38) than for men (19). Words clustered include references to `maternity', `birthmother' (women), and also `paternity' (men). \textit{Arts} is also biased towards women, with 4 clusters for women compared with just 1 cluster for men, and including words such as `sew', `needlework' and `soprano' (women). 
Although not that frequent among the set of the 5000 most biased words in the community, labels related to \emph{science} and \emph{maths} are biased towards men, with only one cluster associated with men but no clusters associated with women. Therefore, this analysis shows that our method, in addition to finding what are the most frequent and pronounced biases in the Google News model (shown in Table \ref{tb:gnews_comparison2}), could also reproduce the biases tested\footnote{Note that previous work tested for arbitrary biases, which were not claimed to be the most frequent or pronounced ones.} by previous work.

%%%%%%%%%%%%%%%%%%%%%%%
%
%  DATASETS 
% 
%%%%%%%%%%%%%%%%%%%%%%%

\section{Reddit Datasets}\label{sec:results}
The Reddit datasets used in the remainder of this paper are presented in Table \ref{tb:datasets}, where \emph{Wpc} means average words per comment, and \emph{Word Density} is the average unique new words per comment.
Data was acquired using the Pushshift data platform \cite{Baumgartner2020}. 
All predefined sets of words used in this work and extended tables are included in Appendixes \ref{ap:sec_extended} and \ref{ap:targetsets}, and the code to process the datasets and embedding models is available publicly\footnote{\url{https://github.com/xfold/LanguageBiasesInReddit}}. We expect to find both different degrees and types of bias and stereotyping in these communities, based on news reporting and our initial explorations of the communities. For instance, /r/TheRedPill and /r/The\_Donald have been widely covered as misogynist and ethnic-biased communities (see below), while /r/atheism is, as far as reporting goes, less biased. 

For each comment in each subreddit, we first preprocess the text by removing special characters, splitting text into sentences, and transforming all words to lowercase. Then, using all comments available in each subreddit and using Gensim word2vec python library, we train a skip-gram word embeddings model of 200 dimensions, discarding all words with less that 10 occurrences (see an analysis varying this frequency parameter in Appendix A) and using a 4 word window. 

\begin{table*}[]
\footnotesize
\centering
\caption{Datasets used in this research}
\begin{tabular}{|l|lrrr|rrr|}
\hline
Subreddit & E.Bias & Years           & Authors    & Comments & Unique Words & Wpc & Word Density\\ \hline
/r/TheRedPill    &  gender & 2012-2018      & 106,161    &  2,844,130    & 59,712  & 52.58   & $3.99\cdot10^{-4}$\\        
/r/DatingAdvice  &  gender  &  2011-2018    & 158,758    &   1,360,397   & 28,583 & 60.22    & $3.48\cdot10^{-4}$\\
/r/Atheism       &  religion & 2008-2009    & 699,994    & 8,668,991     & 81,114 & 38.27    & $2.44\cdot10^{-4} $\\       
/r/The\_Donald   &  ethnicity &  2015-2016  & 240,666     & 13,142,696   & 117,060 & 21.27   & $4.18\cdot10^{-4} $\\\hline
\end{tabular}
\label{tb:datasets}
\end{table*}

After training the models, and by using WEAT \cite{Caliskan2017}, we were able to demonstrate whether our subreddits actually include any of the predefined biases found in previous studies. For instance, by repeating the same gender-related WEAT experiments performed in Section \ref{sec:google_news2} in /r/TheRedPill, it seems that the dataset may be gender-biased, stereotyping men as related to \emph{career} and women to \emph{family} ($p$-value of $0.013$). However, these findings do not agree with other commonly observed gender stereotypes, such as those associating men with \emph{science and math} ($p$-value of $0.411$) and women with \emph{arts} ($p$-value of $0.366$). It seems that, if gender biases are occurring here, they are of a particular kind – underscoring our point that predefined sets of concepts may not always be useful to evaluate biases in online communities.\footnote{Due to technical constraints we limit our analysis to the two major binary gender categories – female and male, or women and men – as represented by the lists of associated words.}

%%%%%%%%%%%%%%%%
%
% THE RED PILL
%
%%%%%%%%%%%%%%%%

\subsection{Gender biases in /r/TheRedPill}\label{sec:trp}
The main subreddit we analyse for gender bias is The Red Pill (/r/TheRedPill). This community defines itself as a forum for the `discussion of sexual strategy in a culture increasingly lacking a positive identity for men’ \cite{Watson2016}, and at the time of writing hosts around 300,000 users. It belongs to the online Manosphere, a loose collection of misogynist movements and communities such as pickup artists, involuntary celibates (`incels’), and Men Going Their Own Way (MGTOW). The name of the subreddit is a reference to the 1999 film The Matrix: `swallowing the red pill,’ in the community’s parlance, signals the acceptance of an alternative social framework in which men, not women, have been structurally disenfranchised in the west.
Within this `masculinist’ belief system, society is ruled by feminine ideas and values, yet this fact is repressed by feminists and politically correct `social justice warriors’. In response, men must protect themselves against a ‘misandrist’ culture and the feminising of society \cite{Marwick2017,LaViolette2019}. Red-pilling has become a more general shorthand for radicalisation, conditioning young men into views of the alt-right \cite{Marwick2017}. 
Our question here is to which extent our approach can help in discovering biased themes and concerns in this community. 

\begin{table}[]
\scriptsize
\caption{Most gender-biased adjectives in /r/TheRedPill.}
\begin{tabular}{| lll|lll |}
\hline
\multicolumn{3}{|c|}{Female}   & \multicolumn{3}{c|}{Male}      \\ \hline
Adjective        & Bias         & Freq (FreqR)           & Adjective             & Bias          & Freq (FreqR) \\ \hline
bumble           & 	 0.245  	&   648 (8778)          & visionary           	 & 0.265  	     &  100 (22815) \\
casual           & 	 0.224  	&  6773 (1834)          & quintessential      	 & 0.245  	     &  219 (15722) \\
flirtatious      & 	 0.205  	&   351 (12305)         & tactician           	 & 0.229   	     &   29 (38426) \\
anal             & 	 0.196  	&  3242 (3185)          & bombastic           	 & 0.199  	     &   41 (33324) \\
okcupid          & 	 0.187  	&  2131 (4219)          & leary               	 & 0.190  	     &   93 (23561) \\
fuckable         & 	 0.187  	&  1152 (6226)          & gurian              	 & 0.185   	     &   16 (48440) \\
unicorn          & 	 0.186  	&  8536 (1541)          & legendary           	 & 0.183  	     &  400 (11481) \\
\hline
\end{tabular}
\label{tb:ktrp}
\end{table}

Table \ref{tb:ktrp} shows the top 7 most gender-biased adjectives for the /r/TheRedPill, as well as their bias value and frequency in the model. 
Notice that most female-biased words are more frequently used than male-biased words, meaning that the community frequently uses that set of words in female-related contexts. Notice also that our POS tagging has erroneously picked up some nouns such as \textit{bumble} (a dating app) and \textit{unicorn} (defined in the subreddit’s glossary as a `Mystical creature that doesn’t fucking exist, aka "The Girl of Your Dreams"').

The most biased adjective towards women is \emph{casual} with a bias of 0.224. That means that the average user of /r/TheRedPill often uses the word \emph{casual} in similar contexts as female-related words, and not so often in similar contexts as male-related words. This makes intuitive sense, as the discourse in /r/TheRedPill revolves around the pursuit of `casual' relationships with women.
For men, some of the most biased adjectives are \emph{quintessential}, \emph{tactician}, \emph{legendary}, and \emph{genious}. 
Some of the most biased words towards women could be categorised as related to externality and physical appearance, such as \textit{flirtatious} and \emph{fuckable}. Conversely, the most biased adjectives for men, such as \textit{visionary} and \textit{tactician}, are internal qualities that refer to strategic game-playing. Men, in other words, are qualified through descriptive adjectives serving as indicators of subjectivity, while women are qualified through evaluative adjectives that render them as objects under masculinist scrutiny.

\paragraph{Categorising Biases}

\begin{table}[tbh]
\scriptsize
\centering
\caption{Comparison of most relevant cluster labels between biased words towards women and men in /r/TheRedPill.}
\begin{tabular}{| l r || r r |} \hline
Cluster Label  & SentW & R. Female & R. Male \\ \hline
\multicolumn{4}{c}{Relevant to \emph{Female}} \\ \hline
Anatomy and physiology                        &  -0.120   & 	1 & 	25              \\
Relationship: Intimate/sexual                 & -0.035    & 	2 & 	30              \\
Judgement of appearance (pretty etc.)         & 0.475    & 	3 & 	40              \\
Evaluation:- Good/bad                         & 0.110    & 	4 & 	2               \\
Appearance and physical properties            & 0.018    & 	10 & 	6               \\ \hline
\multicolumn{4}{c}{Relevant to \emph{Male}} \\ \hline
Power, organizing                             & 0.087    & 	61 & 	1               \\
Evaluation:- Good/bad                         & 0.157    & 	4 & 	2               \\
Education in general                          & 0.002    & 	- & 	4               \\
Egoism                                        & 0.090    & 	- & 	5               \\
Toughness; strong/weak                        & -0.004    & 	- & 	7               \\ \hline
\end{tabular}
\label{tb:trp_comparison}
\end{table}

We now cluster the most-biased words in 45 clusters, using $r=0.15$ (see an analysis of the effect $r$ has in Appendix A), generalising their semantic content.
Importantly, due to this categorisation of biases instead of simply using most-biased words, our method is less prone to stability issues associated with word embeddings \cite{Antoniak2018}, as changes in particular words do not directly affect the overarching concepts explored at the cluster level and the labels that further abstract their meaning (see the stability analysis part in Appendix \ref{ap:sec_granularity}).

Table \ref{tb:trp_comparison} shows some of the most frequent labels for the clusters biased towards women and men in /r/TheRedPill, and compares their importance for each gender.
\emph{SentW} corresponds to the average sentiment of all clusters tagged with the label, as described in Equation \ref{eq:sav}. The \emph{R.Woman} and \emph{R.Male} columns show the rank of the labels for the female and male-biased clusters. \emph{'-'} indicates that no clusters were tagged with that label.

\emph{Anatomy and Physiology}, \emph{Intimate sexual relationships} and \emph{Judgement of appearance} are common labels demonstrating bias towards women in /r/TheRedPill, while the biases towards men are clustered as \emph{Power and organising, Evaluation, Egoism, and toughness}. Sentiment scores indicate that the first two biased clusters towards women carry negative evaluations, whereas most of the clusters related to men contain neutral or positively evaluated words. Interestingly, the most frequent female-biased labels, such as \emph{Anatomy and physiology} and \emph{Relationship: Intimate/sexual} (second most frequent), are only ranked 25th and 30th for men (from a total of 62 male-biased labels). A similar difference is observed when looking at male-biased clusters with the highest rank: \emph{Power, organizing} (ranked 1st for men) is ranked 61st for women, while other labels such as \emph{Egoism} (5th) and \emph{Toughness; strong/weak} (7th), are not even present in female-biased labels.

%%%%%%%%%%%%%%%%
%
% /r/Dating\_Advice
%
%%%%%%%%%%%%%%%%

\paragraph{Comparison to /r/Dating\_Advice}\label{sec:da}
In order to assess to which extent our method can differentiate between more and less biased datasets – and to see whether it picks up on less explicitly biased communities – we compare the previous findings to those of the subreddit /r/dating\_advice, a community with 908,000 members. The subreddit is intended for users to `Share (their) favorite tips, ask for advice, and encourage others about anything dating'. The subreddit's About-section notes that `[t]this is a positive community. Any bashing, hateful attacks, or sexist remarks will be removed', and that `pickup or PUA lingo' is not appreciated. As such, the community shows similarities with /r/TheRedPill in terms of its focus on dating, but the gendered binarism is expected to be less prominently present.

As expected, the bias distribution here is weaker than in /r/TheRedPill. The most biased word towards women in /r/dating\_advice is \emph{floral} with a bias of 0.185, and \emph{molest} (0.174) for men. Based on the distribution of biases (following the method in Section 3.1), we selected the top 200 most biased adjectives towards the `female' and `male' target sets and clustered them using k-means ($r=0.15$), leaving 30 clusters for each target set of words. 
The most biased clusters towards women, such as \emph{(okcupid, bumble)}, and \emph{(exotic)}, are not clearly negatively biased (though we might ask questions about the implied exoticism in the latter term). The biased clusters towards men look more conspicuous: \emph{(poor)}, \emph{(irresponsible, erratic, unreliable, impulsive)} or \emph{(pathetic, stupid, pedantic, sanctimonious, gross, weak, nonsensical, foolish)} are found among the most biased clusters. On top of that, \emph{(abusive)}, \emph{(narcissistic, misogynistic, egotistical, arrogant)}, and \emph{(miserable, depressed)} are among the most sentiment negative clusters. These terms indicate a significant negative bias towards men, evaluating them in terms of unreliability, pettiness and self-importance.

\begin{table}[]
\scriptsize
\centering
\caption{Comparison of most relevant cluster labels between biased words towards women and men in /r/dating\_advice.}
\begin{tabular}{| l r || r r |}
 \hline
Cluster Label  & SentW & R. Female & R. Male  \\ \hline
\multicolumn{4}{c}{Relevant to \emph{Female}} \\ \hline
Quantities                          &   0.202    & 	1 & 	6                       \\
Geographical names                  &   0.026    & 	2 & 	-               \\
Religion and the supernatural       &   0.025    & 	3 & 	-   \\
Language, speech and grammar        &   0.025    & 	4 & 	-       \\
Importance: Important               &   0.227    & 	5 & 	-           \\ \hline
\multicolumn{4}{c}{Relevant to \emph{Female}} \\ \hline
Evaluation:- Good/bad                   &   -0.165    & 	14 & 	1            \\
Judgement of appearance (pretty etc.)   &   -0.148    & 6 & 2   \\
Power, organizing                       &   0.032    & 	51 & 	3                \\
General ethics                          &   -0.089    & 	- & 	4                   \\
Interest/boredom/excited/energetic      &   0.354    & - & 5      \\ \hline
\end{tabular}
\label{tb:da_comparison}
\end{table}

No typical bias can be found among the most common labels for the k-means clusters for women. \emph{Quantities} and \emph{Geographical names} are the most common labels. The most relevant clusters related to men are \emph{Evaluation} and \emph{Judgement of Appearance}, together with \emph{Power, organizing}. Table \ref{tb:da_comparison} compares the importance between some of the most relevant biases for women and men by showing the difference in the bias ranks for both sets of target words. The table shows that there is no physical or sexual stereotyping of women as in /r/TheRedPill, and \emph{Judgment of appearance}, a strongly female-biased label in /r/TheRedPill, is more frequently biased here towards men (rank 2) than women (rank 6). Instead we find that some of the most common labels used to tag the female-biased clusters are \emph{Quantities}, \emph{Language, speech and grammar} or \emph{Religion and the supernatural}. This, in conjunction with the negative sentiment scores for male-biased labels, underscores the point that /r/Dating\_Advice seems slightly biased towards men.

%%%%%%%%%%%%%%%%
%
% ATHEISM
%
%%%%%%%%%%%%%%%%

\subsection{Religion biases in /r/Atheism}\label{sec:atheism}

In this next experiment, we apply our method to discover religion-based biases. The dataset derives from the subreddit /r/atheism, a large community with about 2.5 million members that calls itself `the web's largest atheist forum', on which `[a]ll topics related to atheism, agnosticism and secular living are welcome'. Are monotheistic religions considered as equals here? To discover religion biases, we use target word sets \emph{Islam} and \emph{Christianity} (see Appendix \ref{ap:sec_extended}).

In order to attain a broader picture of the biases related to each of the target sets, we categorise and label the clusters following the steps described in Section \ref{sec:discoveringstereotypes}. Based on the distribution of biases we found here, we select the 300 most biased adjectives and use an $r=0.15$ in order to obtain 45 clusters for both target sets.
We then count and compare all clusters that were tagged with the same label, in order to obtain a more general view of the biases in /r/atheism for words related to the Islam and Christianity target sets.

\begin{table}[]
\scriptsize
\centering
\caption{Comparison of most relevant labels between \emph{Islam} and \emph{Christianity} word sets for /r/atheism}
\begin{tabular}{| l r || r r |}
\hline
Cluster Label & SentW & R. Islam & R. Chr. \\ \hline
\multicolumn{4}{c}{Relevant to \emph{Islam}} \\ \hline
Geographical names                               & 0   & 	1 & 	39       \\
Crime, law and order: Law and order               & -0.085  & 	2 & 	40   \\
Groups and affiliation                          & -0.012 & 3 & 20    \\
Politeness                                      & -0.134 & 	4 & 	-              \\
Calm/Violent/Angry                                & -0.140  & 	5 & 	-   \\
 \hline
\multicolumn{4}{c}{Relevant to \emph{Christianity}} \\ \hline
Religion and the supernatural                      & 0.003         & 	13 & 	1    \\
Time: Beginning and ending                          & 0        & 	- & 	2       \\
Time: Old, new and young; age                       & 0.079        & 	- & 	3   \\
Anatomy and physiology                              & 0        & 	22 & 	4        \\ 
Comparing:- Usual/unusual                           & 0 & 	- & 	5                                           \\
\hline
\end{tabular}
\label{tb:atheism_comparison}
\end{table}

Table \ref{tb:atheism_comparison} shows some of the most common clusters labels attributed to Islam and Christianity (see Appendix \ref{ap:sec_extended} for the full table), and the respective differences between the ranking of these clusters, as well as the average sentiment of all words tagged with each label. The `-' symbol means that a label was not used to tag any cluster of that specific target set. 
Findings indicate that, in contrast with Christianity-biased clusters, some of the most frequent cluster labels biased towards Islam are \emph{Geographical names}, \emph{Crime, law and order} and  \emph{Calm/Violent/Angry}. On the other hand, some of the most biased labels towards Christianity are \emph{Religion and the supernatural}, \emph{Time: Beginning and ending} and \emph{Anatomy and physiology}.

All the mentioned biased labels towards Islam have an average negative polarity, except for \emph{Geographical names}. Labels such as \emph{Crime, law and order} aggregate words with evidently negative connotations such as \emph{uncivilized, misogynistic, terroristic and  antisemitic}. \emph{Judgement of appearance}, \emph{General ethics}, and \emph{Warfare, defence and the army} are also found among the top 10 most frequent labels for Islam, aggregating words such as \emph{oppressive, offensive and  totalitarian} (see  Appendix \ref{ap:sec_extended}).
However, none of these labels are relevant in Christianity-biased clusters. Further, most of the words in Christianity-biased clusters do not carry negative connotations. Words such as \emph{unitarian, presbyterian, episcopalian} or \emph{anglican} are labelled as belonging to \emph{Religion and the supernatural}, \emph{unbaptized} and \emph{eternal} belong to \emph{Time} related labels, and \emph{biological, evolutionary} and \emph{genetic} belong to \emph{Anatomy and physiology}.

Finally, it is important to note that our analysis of conceptual biases is meant to be more suggestive than conclusive, especially on this subreddit in which various religions are discussed, potentially influencing the embedding distributions of certain words and the final discovered sets of conceptual biases. Having said this, and despite the community’s focus on atheism, the results suggest that labels biased towards Islam tend to have a negative polarity when compared with Christian biased clusters, considering the set of 300 most biased words towards Islam and Christianity in this community. Note, however, that this does not mean that those biases are the most frequent, but that they are the most pronounced, so they may be indicative of broader socio-cultural perceptions and stereotypes that characterise the discourse in /r/atheism. Further analysis (including word frequency) would give a more complete view.

%%%%%%%%%%%%%%%%
%
% THE_DONALD
%
%%%%%%%%%%%%%%%%

\subsection{Ethnic biases in /r/The\_Donald}\label{sec:tdonald}
In this third and final experiment we aim to discover ethnic biases. Our dataset was taken from /r/The\_Donald, a subreddit in which participants create discussions and memes supportive of U.S. president Donald Trump. Initially created in June 2015 following the announcement of Trump's presidential campaign, /r/The\_Donald has grown to become one of the most popular communities on Reddit. Within the wider news media, it has been described as hosting conspiracy theories and racist content \cite{Romano2017}.

For this dataset, we use target sets to compare \emph{white last names}, with \emph{Hispanic names}, \emph{Asian names} and \emph{Russian names}  (see  Appendix \ref{ap:targetsets}). The bias distribution for all three tests is similar: the \emph{Hispanic}, \emph{Asian} and \emph{Russian} target sets are associated with stronger biases than the \emph{white names} target sets. The most biased adjectives towards white target sets include \emph{classic}, \emph{moralistic} and \emph{honorable} when compared with all three other targets sets. Words such as \emph{undocumented}, \emph{undeported} and \emph{illegal} are among the most biased words towards Hispanics, while \emph{Chinese} and \emph{international} are among the most biased words towards Asian, and \emph{unrefined} and \emph{venomous} towards Russian. The average sentiment among the most-biased adjectives towards the different targets sets is not significant, except when compared with Hispanic names, i.e. a sentiment of 0.0018 for white names and -0.0432 for Hispanics (p-value of $0.0241$). 

\begin{table}[]
\centering
\scriptsize
\caption{Most relevant labels for Hispanic target set in /r/The\_Donald}
\begin{tabular}{| l r || r r | }
 \hline
Cluster Label & SentW & R. White & R. Hisp.  \\ \hline
Geographical names & 0  & 	- & 	1                       \\
General ethics & -0.349 & 	25 & 	2                        \\
Wanting; planning; choosing & 0 & 	- & 	3               \\
Crime, law and order & -0.119 & 	- & 	4               \\
Gen. appearance, phys. properties & -0.154 & 	21 & 	10   \\ \hline
 
\end{tabular}
\label{tb:thedonals_comparison}
\end{table}

Table \ref{tb:thedonals_comparison} shows the most common labels and average sentiment for clusters biased towards Hispanic names using $r=0.15$ and considering the 300 most biased adjectives, which is the most negative and stereotyped community among the ones we analysed in /r/The\_Donald.
Apart from geographical names, the most interesting labels for Hispanic vis-à-vis white names are \emph{General ethics} (including words such as \textit{abusive, deportable, incestual, unscrupulous, undemocratic}), \textit{Crime, law and order} (including words such as \emph{undocumented, illegal, criminal, unauthorized, unlawful, lawful} and \textit{extrajudicial}), and \emph{General appearance and physical properties} (aggregating words such as \textit{unhealthy, obese} and \textit{unattractive}). All of these labels are notably uncommon among clusters biased towards white names – in fact, \emph{Crime, law and order} and \emph{Wanting; planning; choosing} are not found there at all.

%%%%%%%%%%%%%%%%%%%%%%%
%
%  DISCUSSION
% 
%%%%%%%%%%%%%%%%%%%%%%%

\section{Discussion}\label{sec:discussion}

Considering the radicalisation of interest-based communities outside of mainstream culture \cite{Marwick2017}, the ability to trace linguistic biases on platforms such as Reddit is of importance. Through the use of word embeddings and similarity metrics, which leverage the vocabulary used within specific communities, we are able to discover biased concepts towards different social groups when compared against each other. This allows us to forego using fixed and predefined evaluative terms to define biases, which current approaches rely on. Our approach enables us to evaluate the terms and concepts that are most indicative of biases and, hence, discriminatory processes.

As Victor Hugo pointed out in Les Miserables, slang is the most mutable part of any language: `as it always seeks disguise so soon as it perceives it is understood, it transforms itself.' Biased words take distinct and highly mutable forms per community, and do not always carry inherent negative bias, such as \textit{casual} and \textit{flirtatious} in /r/TheRedPill. Our method is able to trace these words, as they acquire bias when contextualised within particular discourse communities. Further, by discovering and aggregating the most-biased words into more general concepts, we can attain a higher-level understanding of the dispositions of Reddit communities towards protected features such as gender. Our approach can aid the formalisation of biases in these communities, previously proposed by \cite{Caliskan2017,garg2018word}. It also offers robust validity checks when comparing subreddits for biased language, such as done by \cite{LaViolette2019}. Due to its general nature – word embeddings models can be trained on any natural language corpus – our method can complement previous research on ideological orientations and bias in online communities in general.

Quantifying language biases has many advantages \cite{Abebe2019}. As a diagnostic, it can help us to understand and measure social problems with precision and clarity. Explicit, formal definitions can help promote discussions on the vocabularies of bias in online settings. Our approach is intended to trace language in cases where researchers do not know all the specifics of linguistic forms used by a community. For instance, it could be applied by legislators and content moderators of web platforms such as the one we have scrutinised here, in order to discover and trace the severity of bias in different communities. As pernicious bias may indicate instances of hate speech, our method could assist in deciding which kinds of communities do not conform to content policies. Due to its data-driven nature, discovering biases could also be of some assistance to trace so-called `dog-whistling' tactics, which radicalised communities often employ. Such tactics involve coded language which appears to mean one thing to the general population, but has an additional, different, or more specific resonance for a targeted subgroup \cite{Haney-Lopez2015}. 

Of course, without a human in the loop, our approach does not tell us much about \textit{why} certain biases arise, what they mean in context, or how much bias is too much. Approaches such as Critical Discourse Analysis are intended to do just that \cite{LaViolette2019}. In order to provide a more causal explanation of how biases and stereotypes appear in language, and to understand how they function, future work can leverage more recent embedding models in which certain dimensions are designed to capture various aspects of language, such as the polarity of a word or its parts of speech \cite{rothe2016word}, or other types of embeddings such as bidirectional transformers (BERT) \cite{devlin2018bert}. 
Other valuable expansions could include to combine both bias strength and frequency in order to identify not only strongly biased words but also frequently used in the subreddit, extending the set of USAS labels to obtain more specific and accurate labels to define cluster biases, and study community drift to understand how biases change and evolve over time. Moreover, specific ontologies to trace each type of bias with respect to protected attributes could be devised, in order to improve the labelling and characterisation of negative biases and stereotypes.

We view the main contribution of our work as introducing a modular, extensible approach for exploring language biases through the lens of word embeddings. Being able to do so without having to construct a-priori definitions of these biases renders this process more applicable to the dynamic and unpredictable discourses that are proliferating online.

\section*{Acknowledgments}
This  work  was  supported  by  EPSRC  under grant  EP/R033188/1.

\bibliographystyle{aaai} 
\bibliography{refs}

%%%%%%%%%%%%%%%%%%%%%%%%%%%%%%
%
%
%
%
% APPENDIX
%
%
%
%
%
%%%%%%%%%%%%%%%%%%%%%%%%%%%%%%

\begin{appendices}
\section{Further experiments on /r/TheRedPill}\label{ap:sec_granularity}
In this section we perform various analysis on different aspects on /r/TheRedPill subreddit. We analyse the effect of changing parameter $r$ to modify partition granularity, analyse the model stability, and study the performance of two POS taggers on Reddit.

\paragraph{\textbf{Partition Granularity}}
The selection of different $r$ values for the k-means clustering detailed in Section \ref{sec:discoveringstereotypes} directly influences the number of clusters in the resulting partition of biased words. Low values of $r$ result in smaller partitions and hence biases defined by bigger (more general) clusters, while higher values of $r$ result in a higher variety of specific USAS labels allowing a more fine-grained analysis of the community biases at the expense of conciseness.

\begin{figure}[tbh]
  \centering
  \includegraphics[width=1.05\linewidth]{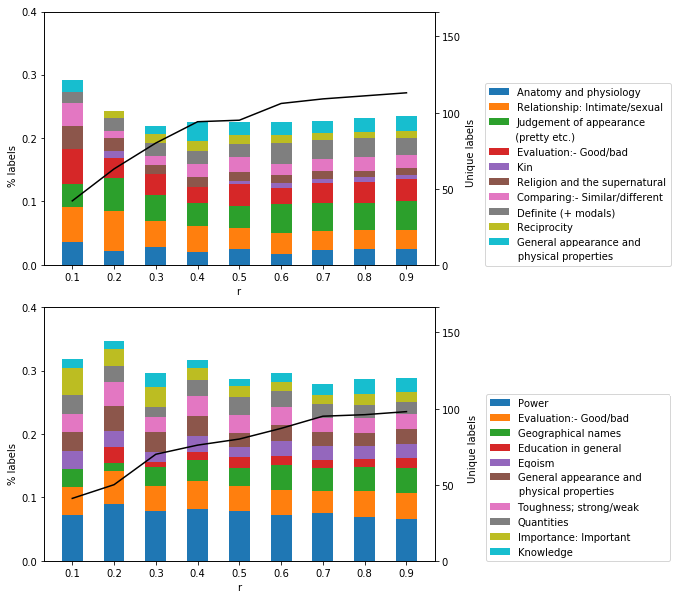}
  \caption{Relative importance (left axis) of the top 10 most frequent labels for women (top) and men (bottom), and number of unique labels (right axis) using different partition granularities ($r$) on /r/TheRedPill}
  \label{fig:app_rfemale}
\end{figure}

Figure \ref{fig:app_rfemale} shows the relative importance of the top 10 most frequent biases in /r/TheRedPill for women and men, presented in section \ref{sec:trp}, together with the quantity of unique USAS labels in each partition obtained for different values of $r$ (see Section \ref{sec:discoveringstereotypes}).
Both figures show that most of the top 10 frequent labels for both women and men (see Section \ref{sec:trp}), have similar relative frequencies when compared with the total of labels in each partition for all values of $r$, with few exceptions such as \emph{Reciprocity} and \emph{Kin} labels for women and \emph{Education in general} for men. This indicates that the set of the most frequent conceptual biases in the community is consistent among different partitions, usually aggregating on average between the 22 and 30\% of the total of the clusters for women and men, despite the increase in the quantity of clusters and unique labels obtained when using higher values of $r$. 
Even considering that the relative frequencies of the presented labels are similar between partitions, the different partitions share, on average, 7 out of the 10 most frequent labels for women and men. Among the top 10 most frequent labels for women in all partitions we find \emph{Anatomy and Physiology}, \emph{Relationship: Intimate/sexual} and \emph{Judgement of appearance}. For men, some of the most frequent labels in all partitions contain \emph{Power}, \emph{Evaluation Good/bad} and \emph{Geographical Names}, among others.

\paragraph{\textbf{Stability analysis}}
To test the stability of our approach we created 10 bootstrapped models of /r/TheRedPill in a similar way as done by \cite{Antoniak2018}, including randomly sampling 50\% of the original dataset and averaging the results over the multiple bootstrapped samples. Results show that the average relative difference between the ranks of the most frequent labels with respect to male and female-related target sets remains similar across all ten sub-datasets. The results show the robustness of our approach to detect conceptual biases, and demonstrated that the biases were extended and shared by the community.

\paragraph{Frequency analysis}
To test the effect that the word frequency threshold (when traininig the word embeddings model) has on the discovered conceptual biases of a community, we trained two new models for /r/TheRedPill changing the minimum frequency threshold to 100 ($f_{100}$) and 1000 ($f_{1000}$). 
First, as a consequence of the increase of the frequency threshold, the new models had relevant vocabulary differences when compared with the original $f_{10}$ presented in Section \ref{sec:trp}: while the original model has a total of 3,329 unique adjectives, $f_{100}$ has 1,540 adjectives (roughly 54\% less), and $f_{1000}$ has 548 adjectives (roughly 84\% less). 
However, a quantitative analysis of the conceptual biases of the models show that the conceptual biases are almost the same for $f_{10}$ and $f_{100}$, and very similar for $f_{1000}$: almost all top 10 labels most biased towards women and men in $f_{10}$, are also biased towards the same target set in $f_{100}$ and $f_{1000}$. The only exception (1 out of the 20 labels for women and men) when comparing $f_{10}$ and $f_{100}$ models is label 'Evaluation:- Good/bad’, a label slightly biased towards men in $f_{10}$ but ranked in the same position for women and men in $f_{100}$. In $f_{1000}$, the figures are very similar too, but as there are many less words in the vocabulary (84\% less), the resulting clusters do not have all labels present in $f_{10}$. However, and very importantly, all labels that do appear in $f_{1000}$ (13 of 20) have the same relative difference as in $f_{10}$, with only two exceptions ('Quantities' and 'Knowledge').

\paragraph{\textbf{Part-of-Speech (POS) analysis in Reddit}}
We performed an experiment comparing the tags provided by the \emph{nltk} POS tagging method with manual annotations performed by us over 100 randomly selected posts from the subreddit /r/TheRedPill, following the same preprocessing presented in the paper, and focusing on adjectives and nouns. The results show that the manual POS tagger agrees with the \emph{nltk} tagger 81.3\% of the times considering nouns over 744 unique nouns gathered from 100 randomly selected comments. For adjectives, the manual tagger agrees with the \emph{nltk} tagger 71.1\% of the times, over 315 unique words tagged as adjectives by any of the two methods (manual, \emph{nltk}) over the same set of comments. In addition, we also compared \emph{nltk} with the \emph{spacy} POS tagger using the same approach.  The results show an agreement of 68.8\% for nouns and 63.7\% for adjectives, obtaining worse results than with the \emph{nltk} library. Although the experiments are not conclusive, (a larger–scale experiment would be needed), the \emph{nltk} library seems to indeed be helpful and be better suited than \emph{spacy} to POS tag on Reddit.

\section{Most Frequent Biased Concepts}\label{ap:sec_extended}
In this section we present the set of top 10 most frequent labels of the communities explored in this work, including all subreddits and Google News. 

\textbf{Top 10 most frequent labels in Google News (Section \ref{sec:google_news2}):} 
{
\textbf{Female}: \emph{Personal names}, \emph{Other proper names}, \emph{Clothes and personal belongings}, \emph{People:- Female}, \emph{Anatomy and physiology}, \emph{Kin}, \emph{Cleaning and personal care}, \emph{Power, organizing}, \emph{Judgement of appearance (pretty etc)}, \emph{medicines and medical treatment}. 
\textbf{Male}: \emph{Personal names}, \emph{Other proper names}, \emph{Warfare}, \emph{Power, organizing}, \emph{Religion and the supernatural}, \emph{Kin},  \emph{Sports}, \emph{Crime}, \emph{Groups and affiliation}, \emph{Games}.                        
}

\textbf{Top 10 most frequent labels in /r/TheRedPill (Section \ref{sec:trp}):} 
{
\textbf{Female}: \emph{Anatomy and physiology}, \emph{Relationship: Intimate/sexual}, \emph{Judgement of appearance (pretty etc.)}, \emph{Evaluation:- Good/bad}, \emph{Kin}, \emph{Religion and the supernatural}, \emph{Comparing:- Similar/different}, \emph{Definite (+ modals)}, \emph{Reciprocity}, \emph{General appearance and physical properties}.
\textbf{Male}: \emph{Power, organizing}, \emph{Evaluation:- Good/bad}, \emph{Geographical names}, \emph{Education in general}, \emph{Egoism}, \emph{General appearance and physical properties}, \emph{Toughness; strong/weak}, \emph{Quantities}, \emph{Importance: Important}, \emph{Knowledge}.       
}

\textbf{Top 10 most frequent labels in /r/Dating\_Advice (Section \ref{sec:da}):} 
{
\textbf{Female}: \emph{Quantities}, \emph{Geographical names}, \emph{Religion and the supernatural}, \emph{Language, speech and grammar}, \emph{Importance: Important}, \emph{Judgement of appearance (pretty etc.)}, \emph{Money: Price}, \emph{Time: Period}, \emph{Science and technology in general}, \emph{Other proper names}.
\textbf{Male}: \emph{Evaluation:- Good/bad}, \emph{Judgement of appearance (pretty etc.)}, \emph{Power, organizing}, \emph{General ethics}, \emph{Interest/boredom/excited/energetic}, \emph{Quantities}, \emph{Happy/sad: Happy}, \emph{General appearance and physical properties}, \emph{Calm/Violent/Angry}, \emph{Helping/hindering}.               
}

\textbf{Top 10 most frequent labels in /r/atheism (Section \ref{sec:atheism}):} 
{
\textbf{Islam}: \emph{Geographical names}, \emph{Crime, law and order: Law and order}, \emph{Groups and affiliation}, \emph{Politeness}, \emph{Calm/Violent/Angry}, \emph{Judgement of appearance (pretty etc.)}, \emph{General ethics}, \emph{Relationship: Intimate/sexual}, \emph{Constraint}, \emph{Warfare, defence and the army; weapons}.
\textbf{Christian}: \emph{Religion and the supernatural}, \emph{Time: Beginning and ending}, \emph{Time: Old, new and young; age}, \emph{Anatomy and physiology}, \emph{Comparing:- Usual/unusual}, \emph{Kin}, \emph{Education in general}, \emph{Getting and giving; possession}, \emph{Time: General: Past}, \emph{Thought, belief}.               
}

\textbf{Top 5 most frequent labels in /r/The\_Donald (Section \ref{sec:tdonald}):} 
{
\textbf{Hispanic}: \emph{Geographical names}, \emph{General ethics}, \emph{Wanting; planning; }, \emph{Crime, law and order}, \emph{Comparing:- Usual/unusual}.
\textbf{Asian}: \emph{Geographical names}, \emph{Government etc.}, \emph{Places}, \emph{Warfare, defence and the army; }, \emph{Groups and affiliation}.
\textbf{Russian}: \emph{Power, organising}, \emph{Quantities}, \emph{Evaluation:- Good/bad}, \emph{Importance: Important}, \emph{Sensory:- Sound}.
}

\section{Target and Evaluative Sets}\label{ap:targetsets}
The sets of words used in this work were taken from \cite{garg2018word}, and \cite{nosek2002harvesting}. For the WEAT test sets performed in Section \ref{sec:google_news2}, we used the same target and attribute word sets used in \cite{Caliskan2017}.
Below, we list all target words sets used.

\textbf{Google News target and attribute sets}\label{sec:gnewssets}
From \cite{garg2018word}.
{
\emph{Female}: sister, female, woman, girl, daughter, she, hers, her. \emph{Male}: brother, male, man, boy, son, he, his, him. \emph{Career words}: executive, management, professional, corporation, salary, office, business, career. \emph{Family}: home, parents, children, family, cousins, marriage, wedding, relatives. \emph{Math}: math, algebra, geometry, calculus, equations, computation, numbers, addition. \emph{Arts}: poetry, art, sculpture, dance, literature, novel, symphony, drama. \emph{Science}: science, technology, physics  , chemistry, Einstein, NASA, experiment, astronomy.
}

\textbf{Google News set of USAS labels related with WEAT experiments:}
{
\emph{Career}: Money \& commerce in industry, Power, organizing. \emph{Family}: Kin, People. \emph{Arts}: Arts and crafts. \emph{Science}: Science and technology in general. \emph{Mathematics}: Mathematics.
}

\textbf{/r/TheRedPill target sets}
From \cite{nosek2002harvesting}.
{
 \emph{Female}: sister, female, woman, girl, daughter, she, hers, her. \emph{Male}: brother, male, man, boy, son, he, his, him.
 }

\textbf{/r/atheism target sets}\label{sec:atheismsets}
From \cite{garg2018word}.
{
 \emph{Islam words:} allah, ramadan, turban, emir, salaam, sunni, koran, imam, sultan, prophet, veil, ayatollah, shiite, mosque, islam, sheik, muslim, muhammad. \emph{Christianity words}: baptism, messiah, catholicism, resurrection, christianity, salvation, protestant, gospel, trinity, jesus, christ, christian, cross, catholic, church
 }

\textbf{r/The\_Donald target sets}\label{sec:tdsets}
From \cite{garg2018word}.
{
 \emph{White last names}: harris, nelson, robinson, thompson, moore, wright, anderson, clark, jackson, taylor, scott, davis,
allen, adams, lewis, williams, jones, wilson, martin, johnson. \emph{Hispanic last names}: ruiz, alvarez, vargas, castillo, gomez, soto, gonzalez, sanchez, rivera, mendoza, martinez, torres, rodriguez, perez, lopez, medina, diaz, garcia, castro, cruz. \emph{Asian last names}: cho, wong, tang, huang, chu, chung, ng, wu, liu, chen, lin, yang, kim, chang, shah, wang, li, khan, singh, hong. \emph{Russian last names}: gurin, minsky, sokolov, markov, maslow, novikoff, mishkin, smirnov, orloff, ivanov, sokoloff, davidoff, savin, romanoff, babinski, sorokin, levin, pavlov, rodin, agin
}

\end{appendices}

\end{document}